\documentclass[iicol,pdflatex,sn-mathphys-num]{sn-jnl}% Math and Physical Sciences Numbered Reference Style 

\usepackage{graphicx}%
\usepackage{multirow}%
\usepackage{amsmath,amssymb,amsfonts}%
\usepackage{amsthm}%
\usepackage{mathrsfs}%
\usepackage[title]{appendix}%
\usepackage{textcomp}%
\usepackage{manyfoot}%
\usepackage{booktabs}%
\usepackage{algorithm}%
\usepackage{algorithmicx}%
\usepackage{algpseudocode}%
\usepackage{listings}%
\usepackage{amsmath}
\usepackage{xspace}
\usepackage{graphicx}
\usepackage{epstopdf}
\usepackage{amsmath}
\usepackage{algorithm}
\usepackage{algorithmicx}
\usepackage{algpseudocode}
\usepackage{url}
\usepackage{hyperref}
\usepackage[table,xcdraw]{xcolor}
\newcommand{\methodname}{MLLM-Selector\xspace}

\newcommand{\addblankpages}[1]{
    \ifnum\value{page}>1\newpage\fi
    \begingroup
    \pagestyle{empty}
    \count0=#1
    \loop
        \ifnum\count0>0
            \null
            \newpage
            \advance\count0 by -1
    \repeat
    \endgroup
}

\theoremstyle{thmstyleone}%
\theoremstyle{thmstyletwo}%

\theoremstyle{thmstylethree}%

\begin{document}

\title[\methodname]{\methodname: Necessity and Diversity-driven High-Value Data Selection for Enhanced Visual Instruction Tuning}


\author[1]{\fnm{Yiwei} \sur{Ma}}\email{yiweima@stu.xmu.edu.cn}
\equalcont{These authors contributed equally to this work.}

\author[2]{\fnm{Guohai} \sur{Xu}}\email{guohai.explorer@gmail.com}
\equalcont{These authors contributed equally to this work.}

\author[1]{\fnm{Xiaoshuai} \sur{Sun}}\email{xssun@xmu.edu.cn}

\author[1]{\fnm{Jiayi} \sur{Ji}}\email{jjyxmu@gmail.com}

\author[2]{\fnm{Jie} \sur{Lou}}
\email{loujie0822@gmail.com}

\author[2]{\fnm{Debing} \sur{Zhang}}
\email{dengyang@xiaohongshu.com}

% \author[3]{\fnm{Tat-Seng} \sur{Chua}}\email{dcscts@nus.edu.sg}

\author[1]{\fnm{Rongrong} \sur{Ji}}\email{rrji@xmu.edu.cn}

\affil[1]{\orgdiv{Key Laboratory of Multimedia Trusted Perception and Efficient Computing}, \orgname{Ministry of Education of China, Xiamen University}, \orgaddress{ \postcode{361005}, \state{Fujian}, \country{P.R. China}}}

\affil[2]{\orgname{Xiaohongshu Inc}, \orgaddress{ \postcode{200025}, \state{Shanghai}, \country{P.R. China}}}

% \affil[3]{\orgname{National University of Singapore}, \orgaddress{  \country{Singapore}}}

% \affil[3]{\orgdiv{Department}, \orgname{Organization}, \orgaddress{\street{Street}, \city{City}, \postcode{610101}, \state{State}, \country{Country}}}


\abstract{

Visual instruction tuning (VIT) has emerged as a crucial technique for enabling multi-modal large language models (MLLMs) to follow user instructions adeptly. 
Yet, a significant gap persists in understanding the attributes of high-quality instruction tuning data and frameworks for its automated selection.
To address this, we introduce \methodname, an automated approach that identifies valuable data for VIT by weighing necessity and diversity.
Our process starts by randomly sampling a subset from the VIT data pool to fine-tune a pretrained model, thus creating a seed model with an initial ability to follow instructions.
Then, leveraging the seed model, we calculate necessity scores for each sample in the VIT data pool to identify samples pivotal for enhancing model performance. 
Our findings underscore the importance of mixing necessity and diversity in data choice, leading to the creation of \methodname, our methodology that fuses necessity scoring with strategic sampling for superior data refinement.
Empirical results indicate that within identical experimental conditions, \methodname surpasses LLaVA-1.5 in some benchmarks with less than 1\% of the data and consistently exceeds performance across all validated benchmarks when using less than 50\%.

}




\maketitle

\section{Introduction}\label{sec:intro}

The advent of large language models (LLMs), such as GPT~\cite{radford2019language,brown2020language,ouyang2022training,achiam2023gpt}, Gemma~\cite{team2024gemma}, LLaMA~\cite{touvron2023llama}, InterLM~\cite{team2023internlm,cai2024internlm2}, and QWen~\cite{bai2023qwen,yang2024qwen2}, has profoundly transformed the field of natural language processing (NLP)~\cite{vaswani2017attention,devlin2018bert,chen2021evaluating}. These models have significantly advanced our ability to understand and generate human-like text, revolutionizing applications across domains such as healthcare, finance, and education. To further extend these capabilities into the realm of visual understanding, a series of pioneering studies~\cite{chen2024internvl,li2024llava,lu2024deepseek,wu2023next} have integrated visual encoders into LLMs, leading to the development of multimodal large language models (MLLMs). Notable examples include Gemini~\cite{team2023gemini}, LLaVA~\cite{liu2024visual,liu2024improved}, and QWen-VL~\cite{bai2023qwenvl}.

\begin{figure}[]
\centering
\includegraphics[width=0.5\textwidth]{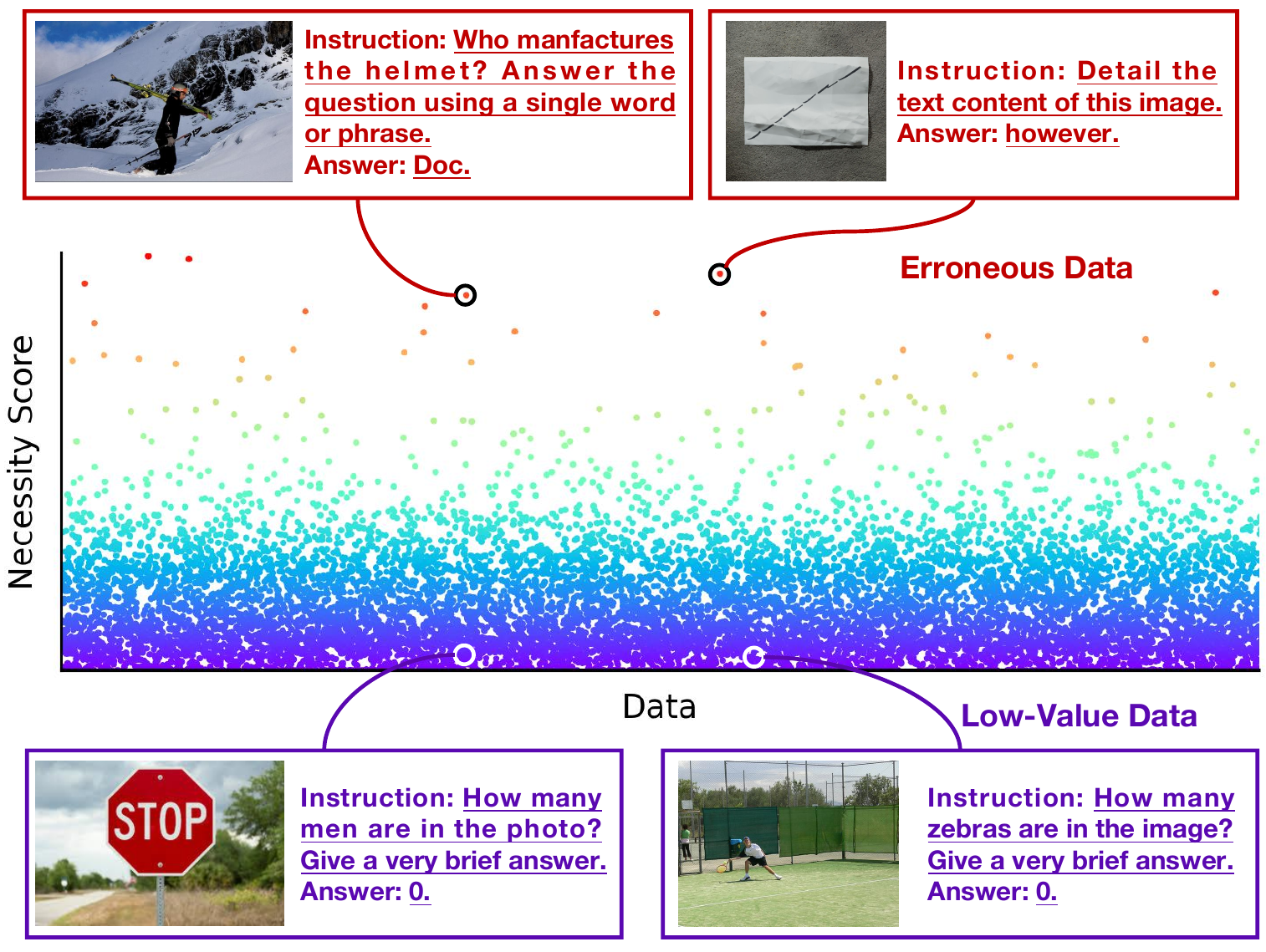} 
\caption{The VIT dataset might include low-value or erroneous data, potentially impairing MLLM performance.}
\label{fig:intro}
\end{figure}

To enhance the instruction-following abilities of MLLMs, visual instruction tuning (VIT)~\cite{liu2024visual,li2023vision,zhu2023minigpt,ma2022xclip} is paramount. This process involves fine-tuning MLLMs with meticulously curated instruction datasets~\cite{chen2023sharegpt4v,liu2024visual}. However, the success of this stage is critically dependent on the quality of VIT data. Utilizing suboptimal data can lead to severe consequences, such as excessive computational resource consumption and diminished model performance due to erroneous or low-value data.
For instance, as illustrated in Figure~\ref{fig:intro}, erroneous data may include incorrectly annotated answers for given images and instructions, which can degrade the model's accuracy and reliability. Similarly, low-value data consists of instructions with low correlation to the relevant image or overly simple instructions. An example would be an instruction asking, ``How many men are in the photo? Give a very brief answer." for an image of a ``stop" sign. Such data does not significantly contribute to enhancing the model's instruction-following capability.
Therefore, constructing high-quality visual instruction datasets is crucial for achieving optimal MLLM performance. 
Current research primarily emphasizes manually curating training datasets from diverse domains or tasks to improve these models' capabilities~\cite{liu2024visual,chen2024internvl,luo2024feast,li2024mini,xu2024llava,zhao2024mg,laurenccon2024matters,ma2023towards,zhang2024tar3d,zhang2025ar,ma2022knowing}. 
While MLLMs fine-tuned on these visual instruction datasets have shown impressive results across numerous downstream multimodal benchmarks, a significant knowledge gap persists, particularly in understanding the defining characteristics of high-quality instruction tuning data and methodologies for its efficient and automated selection.

To address this critical gap, we introduce \methodname—a pioneering approach specifically designed to automatically identify and select high-value data for VIT. 
Specifically, \methodname initiates with random sampling from the VIT data pool to create a seed dataset. This seed dataset forms the foundation for fine-tuning the base model, resulting in a seed model endowed with preliminary instruction-following capabilities. Leveraging the capabilities of this seed model, we compute a necessity score for each sample in the VIT data pool. Samples with higher necessity scores are deemed crucial for further improving model performance, thereby ensuring that the most beneficial data is selected.
Experimental results robustly underscore the importance of maintaining a balance between necessity and diversity in data selection. Consequently, \methodname employs a \textit{{necessity-based grouped sampling strategy}}. This strategy not only focuses on the necessity of the data but also incorporates diversity through strategic grouping. By doing so, it ensures a well-rounded and representative data selection, crucial for comprehensive model training and performance.
To substantiate the efficacy of \methodname, we conducted experiments across a series of benchmarks and compared the results with those of the widely-used open-source MLLM, LLaVA-1.5~\cite{liu2024improved}, under identical experimental settings. Our findings reveal several promising observations:
1) Using less than 1\% of the data (6k vs. 665k), \methodname was able to outperform LLaVA-1.5 on certain benchmarks.
2) With less than 50\% of the data (301k vs. 665k), \methodname exceeded the performance of LLaVA-1.5 on all validated benchmarks.
3) Employing the same amount of data (665k vs. 665k), \methodname achieved substantial improvements on various benchmarks, \emph{e.g.,} +14.54\% on DOCVQA, +25.36\% on ChartQA, and +9.47\% on HallusionBenchmark.

Our contributions are summarized as follows:

\begin{itemize}
    \item We introduce novel criteria for instruction data selection that emphasize necessity and diversity, and demonstrate their pivotal importance through empirical analysis.

    \item We propose a high-value data selection method, \methodname, which comprehensively evaluates the necessity and diversity of each data point to optimize data quality.

    \item Our extensive experimental results validate the effectiveness of \methodname, showing that it can surpass the performance of LLaVA-1.5 with a substantially smaller training dataset, and achieve substantial performance gains when using equivalent amounts of data.
\end{itemize}

\section{Related Wroks}\label{sec:related}

\subsection{Instruction Tuning}

In the past few years, machine learning (ML) and natural language processing (NLP) have experienced tremendous advancement, particularly in the development and training of models. The advent of pre-trained models such as BERT~\cite{devlin2018bert} and GPT~\cite{brown2020language} has revolutionized the field. Fine-tuning these pre-trained models for downstream tasks has become the new standard, leading to substantial performance improvements~\cite{li2020oscar,li2022blip,li2023blip2,li2022mplug,xu2023mplug}.
Specifically, while pre-trained language models possess enormous potential, they are not typically experts in specific domains. To tailor these models for tasks such as sentiment analysis~\cite{medhat2014sentiment,wankhade2022survey,ma2023beat,zhang2021rstnet}, language translation~\cite{dong2015multi,green2013efficacy}, or question answering~\cite{choi2018quac,duan2017question,zhang2025referring,zhang2024temo} on specialized topics, fine-tuning becomes essential. This process involves exposing the model to labeled examples relevant to the targeted tasks, thereby adjusting pretrained parameters and representations to enhance task-specific performance.
Subsequently, the capabilities of larger language models continued to evolve, notably through techniques such as contextual learning using prompts. Recently, a transformative method known as Instruction Tuning (IT)~\cite{zhang2023instruction,longpre2023flan,peng2023instruction} has emerged, making large language models (LLMs) increasingly practical and effective in real-world applications. IT is a critical step in endowing pre-trained foundation models with instruction-following abilities, enabling them to handle complex instructions and respond in a human-like manner. Prominent examples include Alpaca~\cite{taori2023alpaca}, Vicuna~\cite{chiang2023vicuna}, and WizardLM~\cite{xu2023wizardlm}, which have distilled instruction-tuning datasets from GPT-family models.
Building on the concept of IT, visual instruction tuning (VIT)~\cite{zhao2023svit,lin2024comparison,chen2024visual} plays a crucial role in the development of multimodal large language models (MLLMs). VIT aims to equip MLLMs with the ability to follow visual instructions, thus enhancing their versatility and application range. Current research~\cite{instructblip,liu2024improved,ji2022knowing} often involves synthesizing visual instruction data and mixing it with existing academic datasets to form a comprehensive training dataset. MLLMs trained on these mixed datasets have demonstrated exceptional performance across numerous academic benchmarks~\cite{goyal2017making,marino2019ok,schwenk2022okvqa}.
For instance, MiniGPT-4~\cite{zhu2023minigpt} and LLaVA-1.5~\cite{liu2024improved} have organized existing datasets into instruction formats and generated additional instruction data using GPT. Despite these successes, current methods often simply amalgamate all types of instructions, failing to address potential redundancies within and across different task instructions.
Our research introduces an automated method for selecting high-value data for visual instruction tuning, effectively addressing redundancies. This nuanced approach ensures that the models are not only trained on diverse data but are also optimized for quality and relevance, ultimately leading to more robust and efficient instruction-following capabilities.

\subsection{Data Selection for Instruction Tuning}

Redundant and low-quality data in instruction datasets can lead to increased training costs and result in suboptimal model performance, making the selection of high-value data critically important. Recently, the topic of data selection for instruction tuning has gained significant attention in the development of large language models (LLMs). The goal is to curate a subset of data that maximizes model performance while minimizing unnecessary computational expenses.
One pioneering effort in this area is LIMA~\cite{zhou2024lima}, which demonstrated that the vast majority of knowledge within a large model is acquired during the pre-training stage. Consequently, only a limited number of carefully selected instructions are required to fine-tune the model for high-quality content generation. Following this, Alpagasus~\cite{chen2023alpagasus} introduced an innovative automatic filtering approach using ChatGPT to evaluate the quality of each sample, thus significantly improving training efficiency. Instruction Mining~\cite{cao2023instruction} further advanced the field by employing a linear combination of several indicators to assess sample quality, providing a robust framework for data selection.
As data selection becomes increasingly popular in instruction tuning, the approach is also being explored in the context of visual instruction tuning (VIT). VIT aims to enhance multimodal large language models (MLLMs) by enabling them to follow complex visual instructions effectively.
For instance, InstructionGPT-4~\cite{wei2023instructiongpt} employs a set of custom-designed indicators to identify high-quality visual instructions, ensuring the training data is both relevant and valuable. SELE-FILTER~\cite{chen2024your} champions data selection via a trained score network, ensuring robust model performance. Moreover, TIVE~\cite{liu2024less} addresses the challenges of selecting data from highly complex visual instruction mixtures, introducing sophisticated metrics to evaluate each data sample's utility.
However, many of these existing methods rely on additional models or networks for data selection, which can add complexity and computational overhead to the process. Different from these approaches, our proposed \methodname achieves a fully automated data selection process without the need for training supplementary models, such as score networks. This streamlined approach not only simplifies the data selection procedure but also enhances efficiency.

\begin{figure*}[]
\centering
\includegraphics[width=1.0\textwidth]{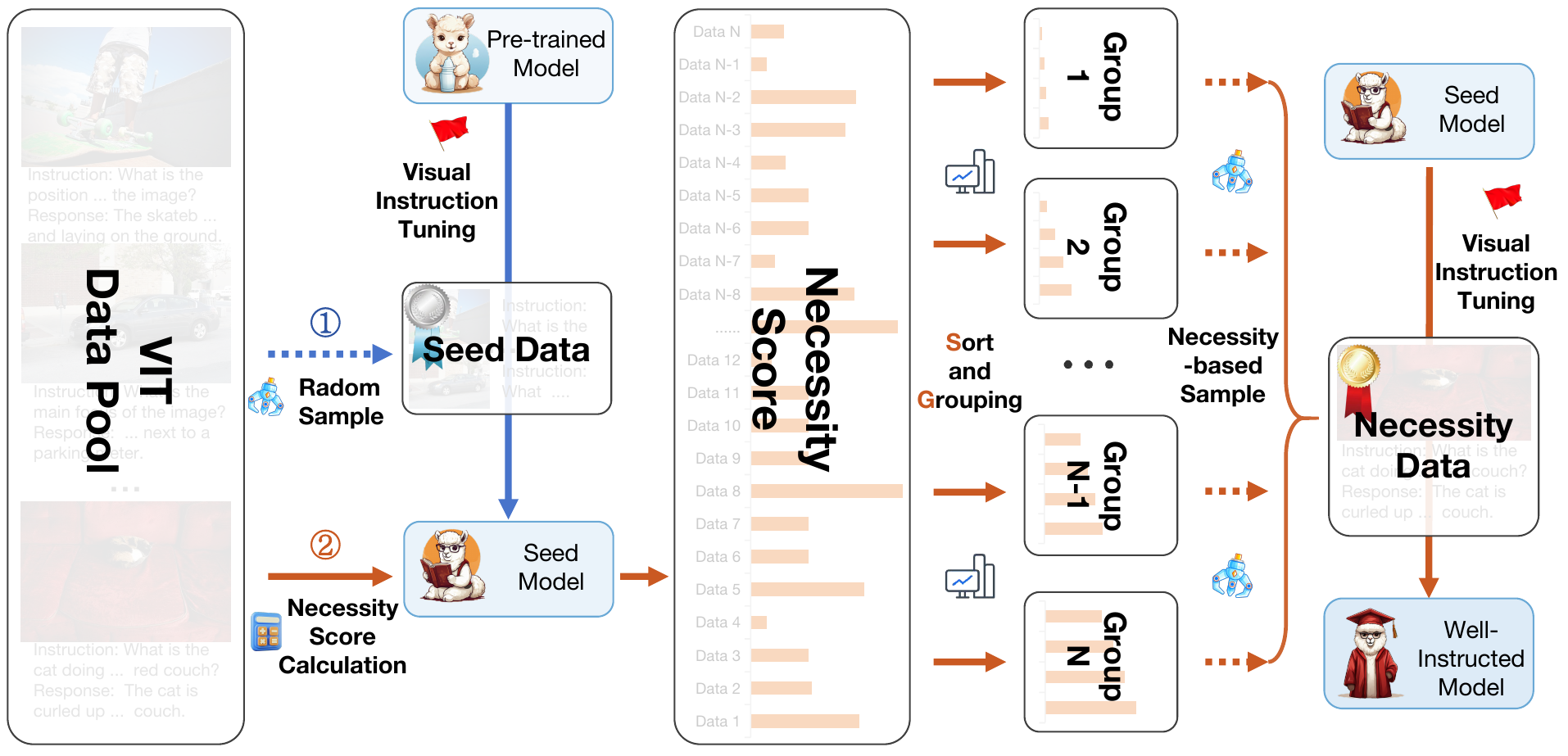} 
\caption{
Overview of \methodname, which consists of two stages: \textit{Stage 1: Initial Seed Data Selection via Random Sampling} and \textit{Stage 2: Necessity Data Selection through Necessity-Based Grouped Sampling}.
% In the first stage, we perform random sampling from the VIT data pool to fine-tune the pre-trained model. 
% This process generates a preliminary model with some instruction-following capabilities, referred to as the seed model.
% In the second stage, the goal is to sample the most essential data for further fine-tuning the seed model. 
% We calculate a necessity score for each sample within the VIT data pool. Using this score, we employ necessity-based grouped sampling to select the most impactful data while also considering diversity. Finally, both the seed data and the necessity data are used to fine-tune the seed model, resulting in a model with enhanced instructional capabilities.
}
\label{fig:overview}
\end{figure*}

\section{Preliminary}

\subsection{Visual Instruction Tuning}
The training process of multi-modal large language models (MLLMs) can be effectively divided into two distinct stages: \textit{Pretraining} and \textit{Visual Instruction Tuning}.
In the \textit{Pretraining} stage, the primary focus is on fine-tuning the projector to harmonize vision and language features. This is typically achieved through the extensive use of image-caption pairs, which serve as a rich source of aligned visual and textual data. By leveraging these pairs, the model learns to intertwine visual inputs with corresponding linguistic descriptions, thereby establishing a foundational understanding of multimodal data. Nevertheless, models resulting from this stage often exhibit limited instruction-following capabilities, which hampers their ability to engage in coherent and meaningful dialogues with human users.
Recognizing the inherent limitations of the pretraining stage, the subsequent \textit{Visual Instruction Tuning}—also referred to as self-supervised fine-tuning—plays a critical role in augmenting the model's interactive capacities. This phase is instrumental in imbuing the pretrained model with enhanced instruction-following abilities. By utilizing a variety of conversation-format data, the model becomes adept at interpreting and responding to nuanced instructions, significantly improving its conversational fluency and responsiveness. 
In this paper, we present a novel approach aimed at automatically selecting high-value visual instruction tuning data to achieve better training efficiency. 
% By focusing on the most informative and relevant data, we aim to reduce training costs while simultaneously improving model performance.

\subsection{Task Formulation}
In this section, we delineate the formulation of the data selection task for visual instruction tuning (VIT), a crucial step towards optimizing the training efficiency and performance of our multi-modal large language models (MLLMs).

Consider a VIT data pool denoted as $\mathcal{D}^M = \{x_i^t, x_i^v\}_{i=1}^M$, where each $(x_i^t, x_i^v)$ pair represents the textual and visual components, respectively, of the $i$-th sample in the data pool, and $M$ represents the total number of samples in this pool. The primary objective of the VIT data selection task is to identify and select a high-value subset $\mathcal{D}^m \subseteq \mathcal{D}^M$, where $m$ is the size of the desired subset. This subset $\mathcal{D}^m$ retains the most informative and relevant examples from the original data pool.

This selective approach is imperative for several reasons. First, utilizing the entire data pool $\mathcal{D}^M$ for fine-tuning can be computationally prohibitive and inefficient, especially as the size of the data pool scales. Second, not all data points contribute equally to the model's learning process; some may be redundant or even detrimental to the model's performance. Therefore, by carefully curating a high-value subset $\mathcal{D}^m$, we can significantly enhance the efficiency of the fine-tuning process.

In this task, the pretrained model $\mathbf{f}^{\text{pre}}$, obtained after the pretraining stage, is subsequently fine-tuned using the selected subset $\mathcal{D}^m$. The goal is to achieve superior performance across a wide range of downstream tasks, leveraging the distilled and more potent training data.

\section{\methodname}

% In this section, we unveil our proposed \methodname, illustrated in Figure~\ref{fig:overview}, an innovative methodology for the automatic selection of high-value data for MLLMs. Initially, we elucidate the design motivation behind \methodname in Sec.~\ref{sec:motivation}. Subsequently, in Sec.~\ref{sec:stage1}, we detail the process of Initial Seed Data Selection via Random Sampling, which serves to impart an initial instruction-following ability to the pretrained model. Finally, in Sec.~\ref{sec:stage2}, we describe the Necessity Data Selection phase, utilizing Necessity-Based Grouped Sampling to systematically identify and rectify gaps within the seed model. 

In this section, we introduce our proposed \methodname, depicted in Figure~\ref{fig:overview}. This innovative methodology automatically selects high-value data for MLLMs. Initially, in Section~\ref{sec:stage1}, we outline the process of initial seed data selection via random sampling, which imparts an initial instruction-following ability to the pretrained model. Subsequently, in Section~\ref{sec:stage2}, we describe the necessity data selection phase, where necessity-based grouped sampling is used to systematically identify and address gaps within the seed model.

\begin{algorithm}
\caption{Stage 1: Initial Seed Data Selection via Random Sampling}\label{alg:seed_data_selection}
\begin{algorithmic}[1]
\Procedure{Seed Data Selection and Fine-tuning}{}
\State \textbf{Input:} Data pool $\mathcal{D}^M$, pretrained model $\mathbf{f}^{\text{pre}}$, seed data size $n_1$
\State \textbf{Output:} Seed model $\mathbf{f}^{\text{seed}}$

\State \textbf{Step 1: Seed Data Sampling}
\State Randomly sample $n_1$ indices $I$ from $\{1, 2, \ldots, M\}$
\State $\mathcal{D}^{n_1} \gets \{(x_i^t, x_i^v) | i \in I\}$

\State \textbf{Step 2: Fine-tune the Pretrained Model}
\State Fine-tune $\mathbf{f}^{\text{pre}}$ with the seed data $\mathcal{D}^{n_1}$ to obtain the seed model $\mathbf{f}^{\text{seed}}$

\State \textbf{Return:} $\mathbf{f}^{\text{seed}}$
\EndProcedure
\end{algorithmic}
\end{algorithm}

\subsection{Stage 1: Initial Seed Data Selection via Random Sampling.}\label{sec:stage1}

In this stage, our objective is to sample a subset $\mathcal{D}^{n_1}$, referred to as seed data, from the entire data pool $\mathcal{D}^M$. The primary purpose of the seed data is to endow the pretrained model $\mathbf{f}^{\text{pre}}$ with an initial instruction-following ability. To achieve this, we employ the simplest random sampling strategy, selecting $n_1$ data points from $\mathcal{D}^M$ to form $\mathcal{D}^{n_1}$, which is formulated as follows:
\begin{equation}
\mathcal{D}^{n_1} \subseteq \mathcal{D}^M = \{(x_i^t, x_i^v) \ | \ i \in I\},
\label{eq:seed_data}
\end{equation}
where $I$ denotes a set of indices randomly selected from $\{1, 2, \ldots, M\}$ with a size of $n_1$:
\begin{equation}
I \sim \text{Uniform}(\{1, 2, \ldots, M\}, n_1).
\label{eq:uniform_sampling}
\end{equation}

Subsequently, we use this seed data $\mathcal{D}^{n_1}$ to fine-tune the pretrained model $\mathbf{f}^{\text{pre}}$, resulting in the seed model $\mathbf{f}^{\text{seed}}$. The procedure for initial seed data selection is summarized in Algorithm \ref{alg:seed_data_selection}.

% %%%%%%%%%%%%%%%%

\begin{algorithm}
\caption{Stage 2: Necessity Data Selection via Necessity-Based Grouped Sampling}\label{alg:necessity_sampling}
\begin{algorithmic}[1]
\Procedure{Necessity Sampling and Fine-tuning}{}
\State \textbf{Input:} Seed model $\mathbf{f}^{\text{seed}}$, data pool $\mathcal{D}^M$, selected data size $n_2$, group size $k$
\State \textbf{Output:} Well-instructed model $\mathbf{f}^{\text{ins}}$

\State \textbf{Step 1: Compute Necessity Scores}
\For{each sample in $\mathcal{D}^M$}
    \State Compute the necessity score $s^{\text{nec}}$ using Equation \eqref{eq:necessity_score}
\EndFor

\State \textbf{Step 2: Sort and Group Samples}
\State Sort samples based on $s^{\text{nec}}$ 
\State Divide the sorted samples into $N$ groups, $\{\mathcal{G}_i\}_{i=1}^{N}$, each containing $k$ samples

\State \textbf{Step 3: Compute Sampling Probabilities}
\For{each group $\mathcal{G}_j$}
    \For{each sample $x_i \in \mathcal{G}_j$}
        \State Compute the sampling probability $p_i$ using Equation \eqref{eq:sampling_probability}
    \EndFor
\EndFor

\State \textbf{Step 4: Sample Data from Each Group}
\For{each group $\mathcal{G}_j$}
    \State Select $\frac{n_2}{N}$ samples based on $p_i$
\EndFor

\State \textbf{Step 5: Aggregate Selected Samples}
\State Merge selected samples from all groups to form $\mathcal{D}^{n_2}$
\State Combine $\mathcal{D}^{n_2}$ with $\mathcal{D}^{n_1}$ to construct $\mathcal{D}^{m}$

\State \textbf{Step 6: Fine-tune the Seed Model}
\State Fine-tune $\mathbf{f}^{\text{seed}}$ using $\mathcal{D}^{m}$

\State \textbf{Return:} Well-instructed model $\mathbf{f}^{\text{ins}}$
\EndProcedure
\end{algorithmic}
\end{algorithm}

\begin{figure*}[]
\centering
\includegraphics[width=1.0\textwidth]{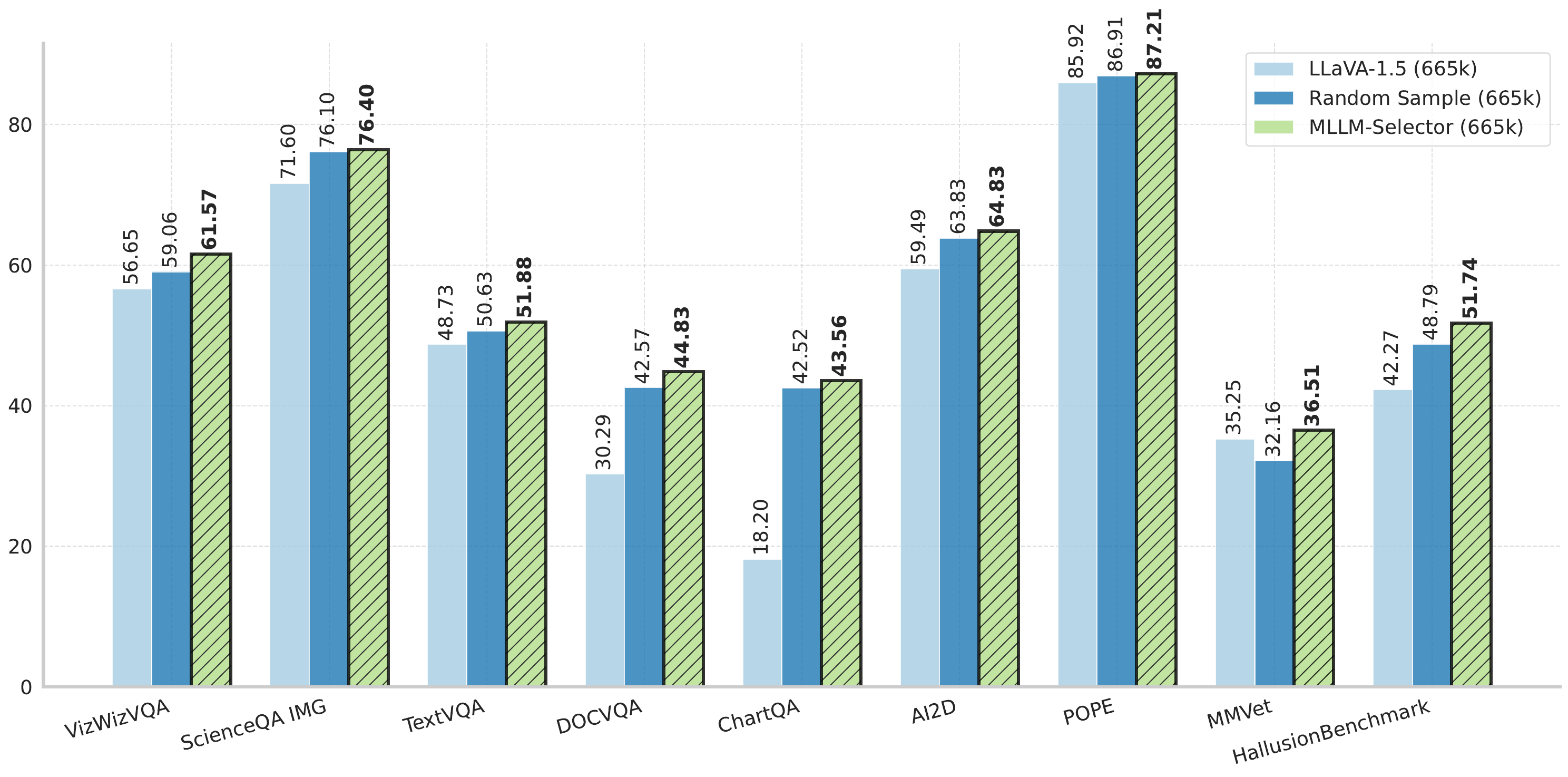} 
% \vspace{-1.5em}
\caption{
Performance comparison of different visual instruction tuning data composition methods. The highest score for each benchmark is highlighted.
}
\label{fig:data_select}
\end{figure*}

\subsection{Stage 2: Necessity Data Selection via Necessity-Based Grouped Sampling}\label{sec:stage2}

In Stage 2, our primary objective is to automatically select the most critical data to optimize the seed model $\mathbf{f}^{\text{seed}}$. For this purpose, we introduce a necessity score, which quantifies the importance of each sample in improving the seed model’s instruction-following capabilities. The necessity score for each sample $(V, T, R_{1:T})$ is defined as:
\begin{equation}
s^{\text{nec}} = \sum_{t=1}^{T} \log \left(p_{\theta} \left( R_{t} \mid V, T, R_{1:t-1} \right) \right),
\label{eq:necessity_score}
\end{equation}
where $V$ represents the visual input, $T$ the instruction, $R_{1:T}$ the ground truth response, and $\theta$ the parameters of the seed model. A higher necessity score $s^{\text{nec}}$ indicates that the sample is more crucial for enhancing the seed model's instruction-following proficiency.
To merely select data with the highest necessity scores directly might seem intuitive, but it often leads to a lack of diversity in the dataset, which hampers overall model performance, as shown in Table~\ref{tab:strategy}. Therefore, we propose the Necessity-Based Grouped Sampling (NBGS) method to balance both necessity and data diversity.
Initially, we sorted the samples based on necessity scores. The sorted samples are then divided into $N$ groups $\{\mathcal{G}_j\}_{j=1}^{N}$, each containing $k$ samples. Within each group $\mathcal{G}_j$, we convert the necessity scores into sampling probabilities using the softmax function:
\begin{equation}
p_i = \frac{\exp(s_i^{\text{nec}})}{\sum_{x_k \in \mathcal{G}_j} \exp(s_k^{\text{nec}})} \quad \text{for each } x_i \in \mathcal{G}_j,
\label{eq:sampling_probability}
\end{equation}
where $s_i^{\text{nec}}$ is the necessity score of $x_i$.
Subsequently, a proportionate number of samples, $\frac{n_2}{N}$, are selected from each group based on the computed sampling probabilities:
\begin{equation}
\mathcal{S}_j = \text{Sample}\left(\mathcal{G}_j, \frac{n_2}{N}, \{p_i \mid x_i \in \mathcal{G}_j\}\right).
\label{eq:group_sampling}
\end{equation}

This balanced sampling strategy ensures that each group effectively contributes to the final dataset, incorporating both high-necessity and diverse samples. The selected samples from each group are then aggregated to form a comprehensive set of necessity data points:
\begin{equation}
\mathcal{D}^{n_2} = \bigcup_{j=1}^{N} \mathcal{S}_j.
\label{eq:final_samples}
\end{equation}

To create the final dataset $\mathcal{D}^{m}$, we merge $\mathcal{D}^{n_2}$ and $\mathcal{D}^{n_1}$, then use it to fine-tune the seed model $\mathbf{f}^{\text{seed}}$ into the well-instructed model $\mathbf{f}^{\text{ins}}$.

\begin{table*}[]
\caption{
Experimental results of different quantities of sampled data using \methodname, with the number of seed data set to 1k. ``See.+Nec." represents the combination of Seed Data + Necessity Data. Performance differences compared to LLaVA-1.5 are highlighted, with values exceeding LLaVA-1.5 in green and those falling short in red.
}
\centering
\setlength{\tabcolsep}{3pt}
\begin{tabular}{ll|cccccc}
\toprule
\textbf{See.+Nec.} & \textbf{Total} & \textbf{ScienceQA} & \textbf{TextVQA} & \textbf{DOCVQA} & \textbf{ChartQA} & \textbf{AI2D}  & \textbf{POPE} \\
\midrule
\rowcolor[HTML]{EFEFEF} 
\textbf{LLaVA-1.5} & \textbf{665k}  & 71.60        & 48.73        & 30.29        & 18.20        & 59.49         & 85.92         \\
\midrule
\textbf{1k+5k}            & \textbf{6k}             & 58.45 \textcolor{red}{\textbf{\tiny{(-13.15)}}}        & 38.04 \textcolor{red}{\textbf{\tiny{(-10.69)}}}        & 26.08 \textcolor{red}{\textbf{\tiny{(-4.21)}}}        & 19.72 \textcolor{green!70!black}{\textbf{\tiny{(+1.52)}}}      & 47.60 \textcolor{red}{\textbf{\tiny{(-11.89)}}}         & 81.09 \textcolor{red}{\textbf{\tiny{(-4.83)}}}         \\

\textbf{1k+10k}           & \textbf{11k}            & 68.67 \textcolor{red}{\textbf{\tiny{(-2.93)}}}        & 42.12 \textcolor{red}{\textbf{\tiny{(-6.61)}}}        & 30.30 \textcolor{green!70!black}{\textbf{\tiny{(+0.01)}}}        & 20.80 \textcolor{green!70!black}{\textbf{\tiny{(+2.60)}}}        & 53.50 \textcolor{red}{\textbf{\tiny{(-5.99)}}}         & 83.98 \textcolor{red}{\textbf{\tiny{(-1.94)}}}         \\
\textbf{1k+50k}           & \textbf{51k}            & 71.69 \textcolor{green!70!black}{\textbf{\tiny{(+0.09)}}}        & 45.08 \textcolor{red}{\textbf{\tiny{(-3.65)}}}        & 35.37 \textcolor{green!70!black}{\textbf{\tiny{(+5.08)}}}        & 23.84 \textcolor{green!70!black}{\textbf{\tiny{(+5.64)}}}        & 57.58 \textcolor{red}{\textbf{\tiny{(-1.91)}}}         & 85.91 \textcolor{red}{\textbf{\tiny{(-0.01)}}}         \\
\textbf{1k+100k}          & \textbf{101k}           & 70.15 \textcolor{red}{\textbf{\tiny{(-1.45)}}}        & 46.53 \textcolor{red}{\textbf{\tiny{(-2.20)}}}        & 36.72 \textcolor{green!70!black}{\textbf{\tiny{(+6.43)}}}        & 27.92 \textcolor{green!70!black}{\textbf{\tiny{(+9.72)}}}        & 57.51 \textcolor{red}{\textbf{\tiny{(-1.98)}}}         & 84.88 \textcolor{red}{\textbf{\tiny{(-1.04)}}}         \\
\textbf{1k+200k}          & \textbf{201k}           & 73.28 \textcolor{green!70!black}{\textbf{\tiny{(+1.68)}}}        & 47.83 \textcolor{red}{\textbf{\tiny{(-0.90)}}}        & 40.51 \textcolor{green!70!black}{\textbf{\tiny{(+10.22)}}}        & 33.72 \textcolor{green!70!black}{\textbf{\tiny{(+15.52)}}}        & 57.29 \textcolor{red}{\textbf{\tiny{(-2.20)}}}        & 86.11 \textcolor{green!70!black}{\textbf{\tiny{(+0.19)}}}  \\
\textbf{1k+300k}          & \textbf{301k}           & 75.21 \textcolor{green!70!black}{\textbf{\tiny{(+3.61)}}}       & 49.46 \textcolor{green!70!black}{\textbf{\tiny{(+0.73)}}}        & 42.06 \textcolor{green!70!black}{\textbf{\tiny{(+11.77)}}}        & 37.88 \textcolor{green!70!black}{\textbf{\tiny{(+19.68)}}}        & 61.08 \textcolor{green!70!black}{\textbf{\tiny{(+1.59)}}}   & 86.46 \textcolor{green!70!black}{\textbf{\tiny{(+0.54)}}}   \\
\textbf{1k+400k}          & \textbf{401k}           & 73.57 \textcolor{green!70!black}{\textbf{\tiny{(+1.97)}}}       & 49.93 \textcolor{green!70!black}{\textbf{\tiny{(+1.20)}}}        & 42.92 \textcolor{green!70!black}{\textbf{\tiny{(+12.63)}}}        & 37.80 \textcolor{green!70!black}{\textbf{\tiny{(+19.60)}}}        & 62.95 \textcolor{green!70!black}{\textbf{\tiny{(+3.46)}}}   & 86.57 \textcolor{green!70!black}{\textbf{\tiny{(+0.65)}}}   \\
\textbf{1k+600k}          & \textbf{601k}           & 76.50 \textcolor{green!70!black}{\textbf{\tiny{(+4.90)}}}       & 50.95 \textcolor{green!70!black}{\textbf{\tiny{(+2.22)}}}        & 43.13 \textcolor{green!70!black}{\textbf{\tiny{(+12.84)}}}        & 39.28 \textcolor{green!70!black}{\textbf{\tiny{(+21.08)}}}        & 63.67 \textcolor{green!70!black}{\textbf{\tiny{(+4.18)}}}   & 86.50 \textcolor{green!70!black}{\textbf{\tiny{(+0.58)}}}  \\

\bottomrule
\end{tabular}
\label{tab:data_amount}
\end{table*}

\section{Experiments}

\subsection{Implementation Details}
To rigorously assess the effectiveness of the proposed \methodname, we conducted experimental validation using LLaVA-1.5~\cite{liu2024improved}, a well-recognized open-source MLLM. LLaVA-1.5 includes 665k visual instruction tuning training samples and comprises a vision encoder, a projector, and an LLM. 
Our methodology adheres strictly to the original instruction-tuning settings and hyperparameters as specified in the official implementations. The experiments were executed on 8 A800 GPUs. 
$k$ is set to 50,000. Unless otherwise specified, the seed and necessity data volumes are 100k and 565k, respectively. The default LLM is Vicuna-13B.
To evaluate performance across different benchmarks, we used the LMMs-Eval toolkit~\cite{lmms_eval2024} for comprehensive evaluation.
% 
% To ensure a fair comparison, we initialized our model using the pre-trained weights of LLaVA-1.5. 
% 
Detailed information regarding the data pool can be found in the Appendix~\ref{sec:data_pool}.

% \subsection{Evaluations}
% We evaluate our proposed methods on a lot of benchmarks

\subsection{Quantitative Experiment}

\subsubsection{What Are the Impacts of Different Selection Strategies?}

To rigorously validate the effectiveness of \methodname, we present a comparative analysis with two representative approaches.
The first approach utilizes a manually constructed VIT dataset, referred to as LLaVA-1.5-mix665k~\cite{liu2024improved}. This dataset, created by human experts leveraging their knowledge and experience, exemplifies a method of manual dataset construction.
To ensure a fair comparison, our automated construction method employs the same dataset size, i.e., 665k samples.
The second approach serves as the baseline for automating VIT dataset construction, where 665k samples are randomly selected from the data pool.
The performance comparison is illustrated in Figure~\ref{fig:data_select}. Our observations are as follows:
\begin{itemize}
    \item Across all nine benchmarks, \methodname consistently outperforms both the manually constructed VIT dataset and the random sampling strategy. This highlights the efficacy of the proposed Necessity-Based Grouped Sampling strategy.
    
    \item In specific benchmarks, such as DOCVQA and ChartVQA, both the random sampling strategy and \methodname substantially surpass the manually constructed VIT dataset. This could be attributed to the diverse nature of the data pool, wherein the sampled data includes a wide range of instructional content related to documents and charts. In contrast, the manually constructed dataset might have overlooked these critical aspects.
\end{itemize}

\subsubsection{How Much Data is Needed to Exceed LLaVA-1.5?}

Previous experiments have established that using 665k data sampled via \methodname results in consistent and significant performance improvements over LLaVA-1.5 under identical experimental settings. In this section, we delve into determining the minimum amount of data that \methodname needs to outperform LLaVA-1.5, while keeping all other settings constant.
We conducted experiments with the seed data fixed at 1k. As shown in Table~\ref{tab:data_amount}, our analyses reveal that on certain datasets, including ScienceQA, DOCVQA, and ChartQA, \methodname requires only 7.7\% of the data (i.e., 1k+50k) to surpass the performance of LLaVA-1.5.
Moreover, our results indicate that \methodname achieves superior performance across all datasets with less over 50\% of the total data volume. Specifically, with 301k total data (1k seed data + 300k necessity data), \methodname consistently outperforms LLaVA-1.5 on all benchmarks. This highlights the efficiency and effectiveness of our data selection method.
For instance, in the case of ScienceQA, our method outperforms LLaVA-1.5 by a margin of 3.61 points with just 301k total data. Similarly, for DOCVQA and ChartQA, our approach shows improvements of 11.77 and 19.68 points, respectively, under the same data constraints. These substantial gains underscore the robustness of our methodology in enhancing performance with a significantly reduced dataset size.

\subsection{Ablation Study}

\begin{figure*}[]
\centering
\includegraphics[width=0.9\textwidth]{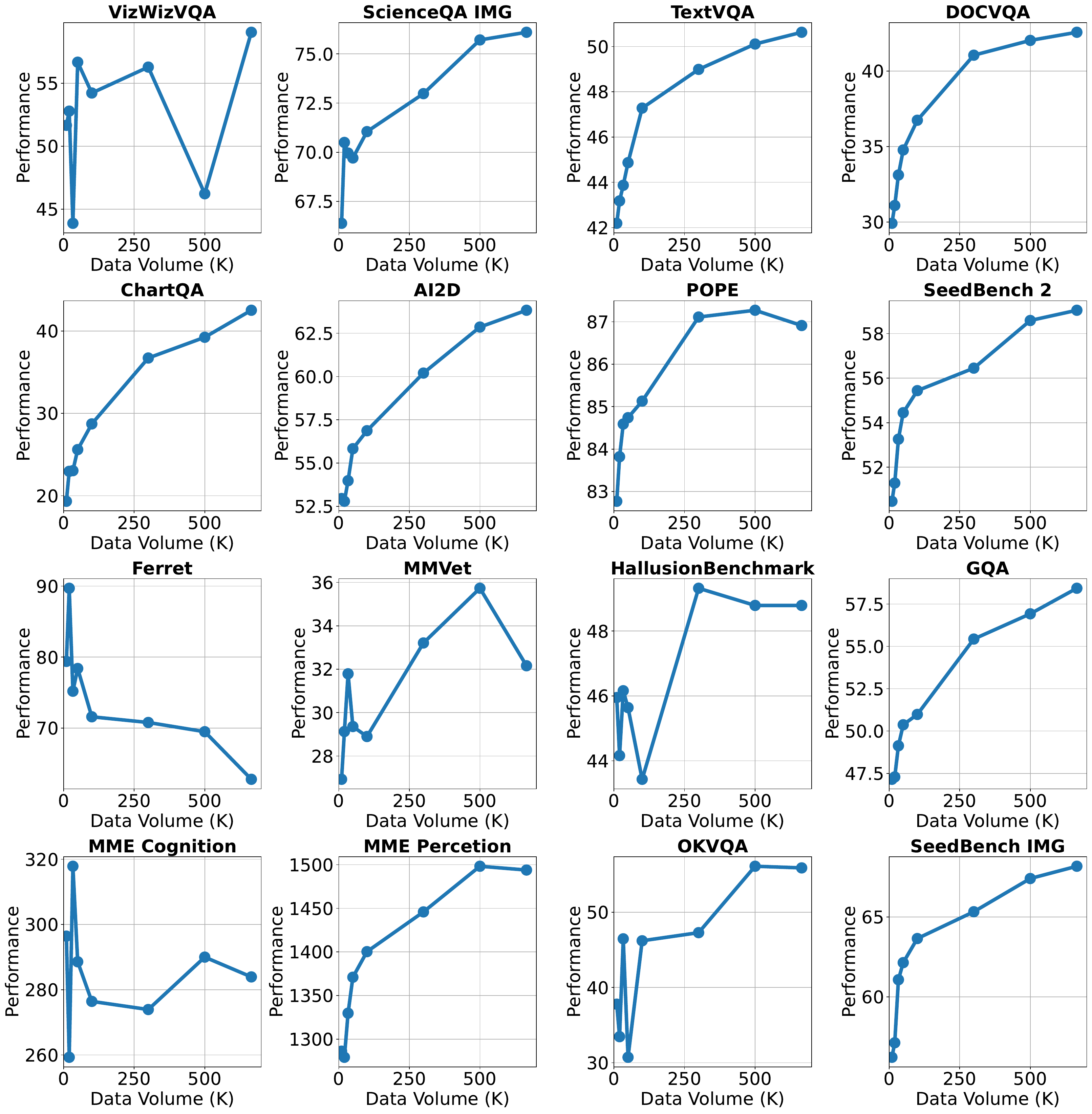} 
\caption{
The relationship between data volume and performance across various benchmarks: VizWizVQA~\cite{gurari2018vizwiz}, ScienceQA IMG~\cite{lu2022learn}, TextVQA~\cite{singh2019towards}, DOCVQA~\cite{mathew2021docvqa}, ChartQA~\cite{masry2022chartqa}, AI2D~\cite{kembhavi2016diagram}, POPE~\cite{li2023evaluating}, SeedBench 2~\cite{li2023seed2}, Ferret~\cite{you2023ferret}, MMVet~\cite{yu2023mm}, HallusionBenchmark~\cite{guan2024hallusionbench}, GQA~\cite{hudson2019gqa}, MME Cognition~\cite{fu2023mme}, MME Perception~\cite{fu2023mme}, OKVQA~\cite{marino2019ok}, and SeedBench IMG~\cite{li2023seed}. 
The data shows how performance metrics change as the data volume increases from 0 to 665 thousand samples for each benchmark.
}
\label{fig:data_volumn}
\end{figure*}

\begin{table}[]
\caption{Performance comparison of Stage 2 with and without checkpoint initialization from Stage 1. SQA, TQA, DQA, and CQA refer to ScienceQA IMG, TextVQA, DOCVQA, and ChartQA, respectively. The best performances are highlighted in bold.}
\setlength{\tabcolsep}{4pt}
\begin{tabular}{l|cccccc}
\toprule
\textbf{Init}         & \textbf{SQA} & \textbf{TQA} & \textbf{DQA} & \textbf{CQA} & \textbf{AI2D}  & \textbf{POPE}  \\
\midrule
 & 76.30                  & 51.33            & \textbf{44.85}  & 42.36            & 64.22          & 86.97          \\
\checkmark   & \textbf{76.40}         & \textbf{51.88}   & 44.83           & \textbf{43.56}   & \textbf{64.83} & \textbf{87.21} \\
\bottomrule
\end{tabular}
\label{tab:init}
\end{table}

\subsubsection{What is the relationship between performance and data volume?}

To investigate the relationship between data volume and model performance across various benchmarks, we conducted a comprehensive analysis by randomly sampling different volumes of training data (\emph{i.e.,} 10k, 20k, 33k, 50k, 100k, 500k, 665k) from our constructed data pool. As illustrated in Figure~\ref{fig:data_volumn}, while most benchmarks exhibit performance improvement as the training data volume increases, some benchmarks do not follow this trend uniformly. Our analysis identifies four distinct patterns in the performance data.
\begin{itemize}
    \item First, we observe benchmarks where performance consistently improves with increasing training data volume. Examples include TextVQA, DOCVQA, ChartQA, AI2D, SeedBench 2, GQA, and SeedBench IMG. These benchmarks indicate a strong data dependency, with models continuing to benefit from additional data up to the maximum tested volume of 665k, without showing signs of saturation.
    \item Second, there are benchmarks where performance initially improves with increasing data volume but reaches saturation before the maximum volume. Notably, POPE and MME Perception achieve peak performance at around 500k data samples. Beyond this volume, further increases in data do not yield performance gains and sometimes even lead to performance decline.
    \item The third pattern involves benchmarks where performance improves overall with increased data volume but exhibits variability, or ``performance jitter," at certain points. ScienceQA IMG, MMVet, HallusionBenchmark, and OKVQA fall into this category. These benchmarks reflect inconsistent performance improvements, suggesting possible inefficiencies or noise in data assimilation.
    \item Lastly, there are benchmarks where model performance shows little to no correlation with the amount of training data. VizWizVQA, Ferret, and MME Cognition exemplify this pattern. Here, additional training data does not translate into significant performance improvements, indicating that other factors may play a more critical role in model efficacy.

\end{itemize}

\begin{table}[]
\caption{Performance comparison of different data selection methods based on necessity score. The best performance for each benchmark is highlighted in bold.}
\setlength{\tabcolsep}{3pt}
\begin{tabular}{l|cccccc}
\toprule
\textbf{Method}           & \textbf{SQA}   & \textbf{TQA}   & \textbf{DQA}   & \textbf{CQA}   & \textbf{AI2D}  & \textbf{POPE}  \\
\midrule
\textbf{Top}    & 75.34          & 51.63          & 43.47          & 41.88          & 64.05          & 86.79          \\
\textbf{Bottom} & 75.53          & 50.79          & 43.58          & 42.76          & 63.90          & 86.59          \\
\textbf{Ours}       & \textbf{76.40} & \textbf{51.88} & \textbf{44.83} & \textbf{43.56} & \textbf{64.83} & \textbf{87.21} \\
\bottomrule
\end{tabular}
\label{tab:strategy}
\end{table}

\begin{table}[]
\caption{
Performance comparison of different $\tau$ values for data selection based on necessity scores. The highest performance for each benchmark is highlighted in bold.
}
\setlength{\tabcolsep}{4pt}
\begin{tabular}{l|cccccc}
\toprule
\textbf{$\tau$} & \textbf{SQA}   & \textbf{TQA}   & \textbf{DQA}   & \textbf{CQA}   & \textbf{AI2D}  & \textbf{POPE}  \\
\midrule
\textbf{0.5}    & 74.67          & 50.46          & 43.77          & 40.44          & 64.77          & 86.83          \\
\textbf{1}      & 76.40          & 51.88          & \textbf{44.83} & \textbf{43.56} & \textbf{64.83} & \textbf{87.21} \\
\textbf{2}      & 75.95          & 51.90          & 43.67          & 41.08          & 64.05          & 86.61          \\
\textbf{10}     & \textbf{76.90} & \textbf{52.06} & 44.48          & 42.32          & 64.51          & 87.06          \\
\bottomrule
\end{tabular}
\label{tab:tau}
\end{table}

\subsubsection{Does Stage 2 Training Require Initialization?}

In this section, we investigate the necessity of using the weights from the seed model trained in Stage 1 as initialization for Stage 2. Table~\ref{tab:init} presents the performance of models trained with and without initialization from Stage 1.
Our findings indicate that initializing Stage 2 training with the checkpoint from Stage 1 generally leads to performance improvements across most benchmarks. This enhancement is likely due to the fact that the necessity data was selected based on the Stage 1 checkpoint. By initializing the model with the seed checkpoint, the selected necessity data appears to have a greater positive impact, as it aligns with the knowledge embedded in the seed model.
Given these observations, we recommend using checkpoint initialization from Stage 1 to enhance Stage 2 training outcomes.

% Stage 1 aims to train a seed model to select necessity data. In this section, we aim to explore whether we need to use the weights of the seed model trained in the first stage as initialization for stage 2.
% Thus,  We perform and do not initialize separately, and display the results in Table~\ref{tab:init}.
% We observe that most benchmarks have achieved performance improvements if using weight initialization.
% This may be because the necessity data is selected based on the stage 1 checkpoint. Thus, if we initliaze the model using seed checkpoint, the necessity data will  have the greatest impact, because these necessity data is selected based on the knowledge of seed model.

\begin{table}[]
\caption{
Performance comparison of different $k$ values for data selection based on necessity scores. The highest performance for each benchmark is highlighted in bold.
}
\setlength{\tabcolsep}{4pt}
\begin{tabular}{l|cccccc}
\toprule
\textbf{$k$} & \textbf{SQA}   & \textbf{TQA}   & \textbf{DQA}   & \textbf{CQA}   & \textbf{AI2D}  & \textbf{POPE}  \\
\midrule
\textbf{1k}     & 77.84          & 51.05          & 44.26          & 42.48          & 65.29          & 86.77          \\
\textbf{5k}     & \textbf{77.94} & 50.69          & 43.04          & 41.60          & 65.19          & 86.52          \\
\textbf{10k}    & 77.49          & 51.55          & 44.40          & 42.24          & \textbf{65.54} & 86.84          \\
\textbf{50k}    & 76.40          & \textbf{51.88} & 44.83          & \textbf{43.56} & 64.83          & 87.21          \\
\textbf{100k}   & 77.44          & 51.49          & 44.25          & 41.92          & 64.41          & 87.01          \\
\textbf{200k}   & 77.44          & 50.77          & \textbf{45.02} & 41.52          & 63.12          & \textbf{87.57} \\
\textbf{500k}   & 77.39          & 51.67          & 43.69          & 41.36          & 64.12          & 87.06          \\
\textbf{1000k}  & 76.40          & 49.83          & 43.34          & 42.76          & 65.09          & 87.19          \\
\bottomrule
\end{tabular}
\label{tab:k}
\end{table}

\begin{table*}[]
\caption{
Experimental results for different distributions of seed data and necessity data quantities. ``See.+Nec." denotes Seed Data + Necessity Data. The total dataset size is consistently maintained at 665k for fair comparison. The best performance for each benchmark is highlighted in bold.
}
\centering
\begin{tabular}{ll|cccccc}
\toprule
\textbf{See.+Nec.} & \textbf{Total} & \textbf{ScienceQA} & \textbf{TextVQA} & \textbf{DOCVQA} & \textbf{ChartQA} & \textbf{AI2D}  & \textbf{POPE} \\
\midrule
\textbf{1k+664k}   & \textbf{665k}           & 77.19                  & 51.49            & 43.78           & 40.92            & 63.92          & 87.20 \\
\textbf{10k+655k}  & \textbf{665k}           & 75.16                  & 51.35            & 44.59           & 41.48            & 64.31          & \textbf{87.24} \\
\textbf{20k+645k}  & \textbf{665k}           & 77.54                  & 50.46            & 44.42           & 42.40            & 64.90          & 86.50 \\
\textbf{33k+632k}  & \textbf{665k}           & 74.62                  & 51.01            & 43.56           & 42.04            & 64.18          & 86.51 \\
\textbf{50k+615k}  & \textbf{665k}           & \textbf{77.74}         & 51.07            & 44.21           & 41.64            & 65.38          & 87.20 \\
\textbf{100k+565k} & \textbf{665k}           & 76.40                  & \textbf{51.88}   & \textbf{44.83}  & \textbf{43.56}   & 64.83          & 87.21 \\
\textbf{300k+365k} & \textbf{665k}           & 76.60                  & 50.51            & 44.51           & 42.80            & \textbf{65.64} & 86.83 \\
\textbf{500k+165k} & \textbf{665k}           & 76.25                  & 51.23            & 44.46           & 43.04            & 65.32          & 86.98 \\
\bottomrule
\end{tabular}
\label{tab:distribution}
\end{table*}

\begin{table*}[]
\setlength{\tabcolsep}{3pt}
\caption{
Performance comparison of different LLM sizes and data construction methods. ``Artificial" represents the manually curated dataset in LLaVA-1.5, ``Random" represents random sampling from our proposed dataset pool, and ``MLLM-Selector" represents our proposed method. The total dataset size is maintained at 665k for fair comparison. The best performance for each benchmark is highlighted in bold.
}
\centering
\begin{tabular}{ll|cccccc}
\toprule
\textbf{LLM Size} & \textbf{Data}   & \textbf{ScienceQA} & \textbf{TextVQA} & \textbf{DOCVQA} & \textbf{ChartQA} & \textbf{AI2D} & \textbf{POPE} \\
\midrule
\multirow{3}{*}{\textbf{\begin{tabular}[c]{@{}l@{}}Vicuna\\ -7B\end{tabular}}}
& \textbf{Artificial}    & 70.43              & 46.07            & 28.08           & 18.24            & 54.79          & 85.87 \\
& \textbf{Random}        & 74.57              & 49.51            & 40.14           & 34.84            & 60.14          & 86.43 \\
& \textbf{MLLM-Selector} & \textbf{75.71}     & \textbf{49.08}   & \textbf{42.0}   & \textbf{37.20}   & \textbf{61.08} & \textbf{87.07} \\
\midrule
\multirow{3}{*}{\textbf{\begin{tabular}[c]{@{}l@{}}Vicuna\\ -13B\end{tabular}}}
& \textbf{Artificial}    & 71.60              & 48.73            & 30.29           & 18.20            & 59.49          & 85.92 \\
& \textbf{Random}        & 76.10              & 50.63            & 42.57           & 42.52            & 63.83          & 86.91 \\
& \textbf{MLLM-Selector} & \textbf{76.40}     & \textbf{51.88}   & \textbf{44.83}  & \textbf{43.56}   & \textbf{64.83} & \textbf{87.21} \\
\bottomrule
\end{tabular}
\label{tab:size}
\end{table*}

\subsubsection{How to Sample Data Based on Necessity Score?}

Once we have computed the necessity score for each sample, it is crucial to adopt effective strategies to select data for training. In this study, we evaluate three distinct strategies to gauge their impact on model performance.
1. \textit{Top Necessity Scores}: We select the top 565k samples with the highest necessity scores, representing the most essential data for the seed model.
2. \textit{Bottom Necessity Scores}: Conversely, we select the 565k samples with the lowest necessity scores, representing the least necessity data for the seed model.
3. \textit{Proposed Method}: Our proposed approach (\methodname) selects data by balancing both necessity and diversity to maximize overall model performance.
The performance results of these three strategies are presented in Table~\ref{tab:strategy}.
The results clearly demonstrate that our proposed method outperforms both the top and bottom necessity score strategies across all benchmarks. This superiority likely stems from our method's ability to consider data diversity, which prevents the model from overfitting to a narrow data distribution and ensures comprehensive learning.
Additionally, the top necessity score strategy does not consistently outperform the bottom necessity score strategy. This inconsistency could be attributed to the presence of excessively difficult samples within the high necessity score group, which may hinder the model's performance improvement. Such challenging samples might overwhelm the model, leading to suboptimal learning outcomes.

% After getting the necessity score for each sample, we can use different strategies to get the data for training.
% Specifically, we leverage three strategies to explore this effect.
% First, we leverage the 565k data with the top necessity scores. This data is the most necessity data for the seed model.
% Second, we leverage the 565k data with the bottom necessity scores. These data are the least necessary for the seed model.
% Finally, we leverage the proposed \methodname to sample data considering both necessity and diversity.
% % 
% The results are displayed in Table~\ref{sec:strategy}.
% % 
% We observe that both the top strategy and bottom strategy perform worse than the \methodname. This may be because that the top strategy and bottom strategy didn't consider diversity.
% % 
% Additionally, we observed the top strategy did not consistently perform better than the bottom strategy. This may be because that the top strategy may contains some too difficult samples for seed model. This may affect the performance improvement of the model.

\begin{figure*}[]
\centering
\includegraphics[width=1.0\textwidth]{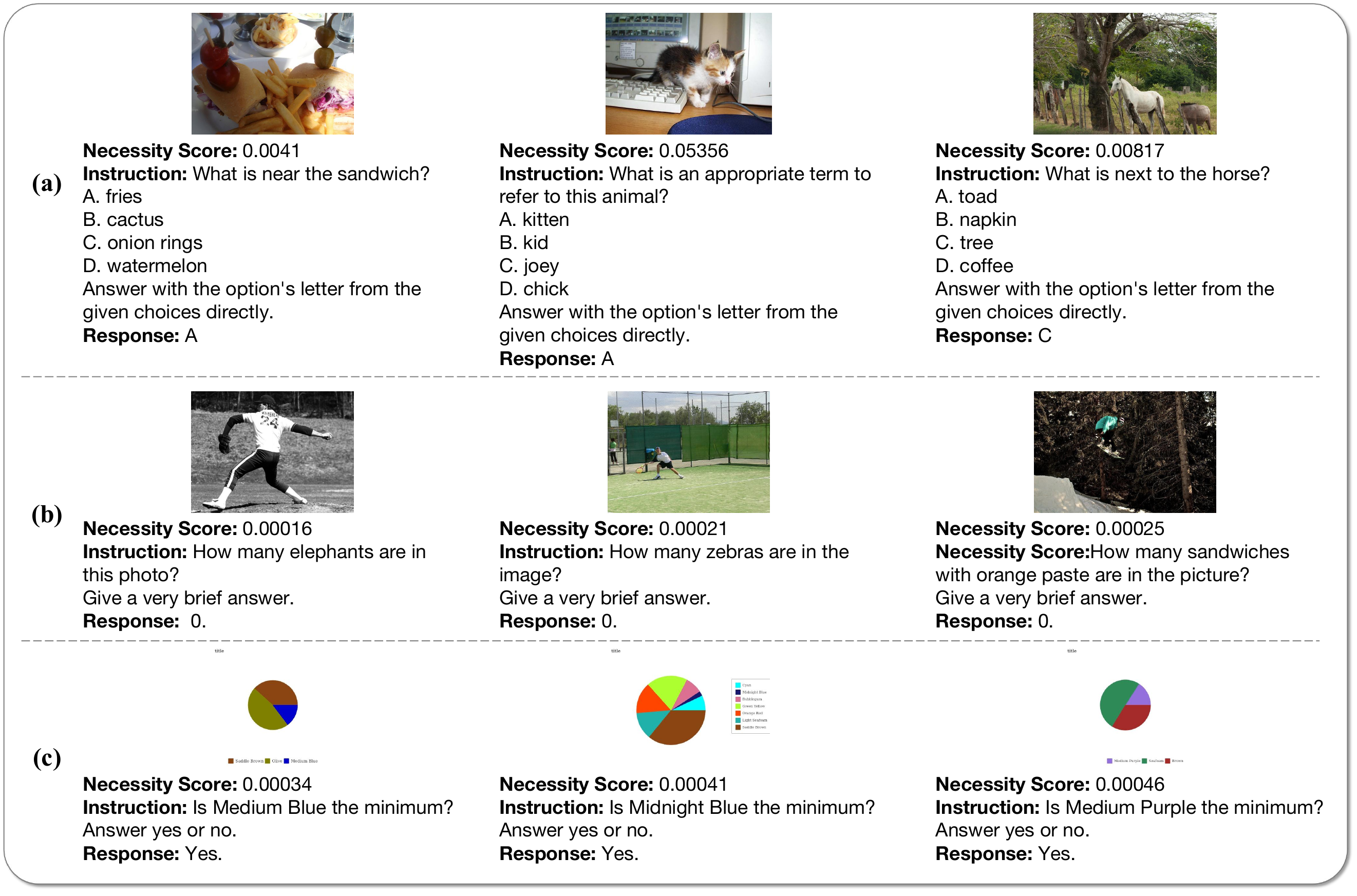} 
\caption{
Examples of samples with low necessity scores, categorized into three types:
(a) Samples with noticeable differences in options, leading to easily solvable problems.
(b) Samples where the problem is loosely related to the image, resulting in straightforward answers.
(c) Samples with overly simple questions about charts.
}
\label{fig:low_score}
\end{figure*}

\begin{figure*}[]
\centering
\includegraphics[width=1.0\textwidth]{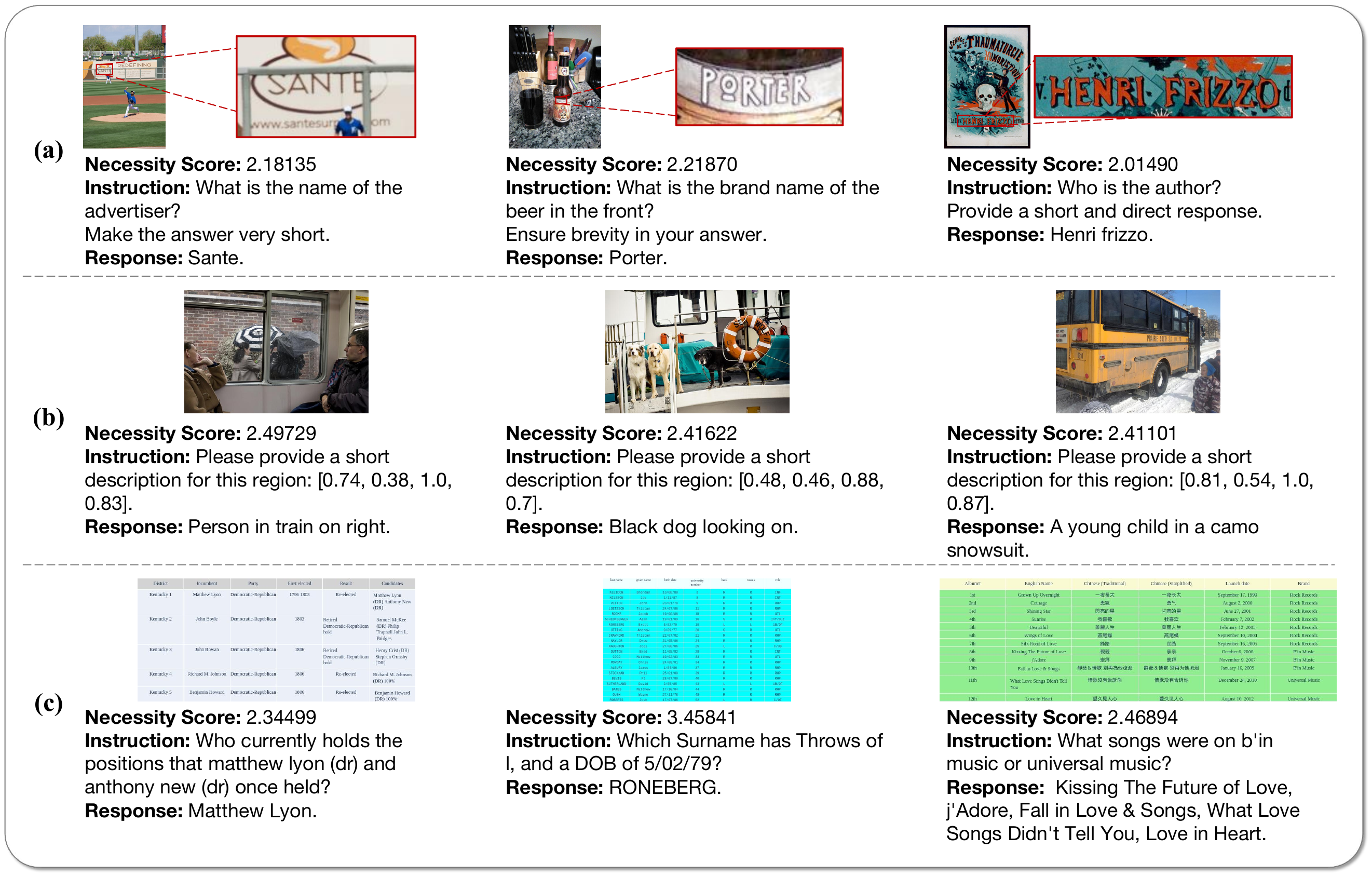} 
\caption{
Examples of samples with high necessity scores, categorized into three types:
(a) Fine-grained text recognition problems, such as identifying specific text in an image.
(b) Understanding specific regions within an image rather than the whole scene, including providing descriptions for specified coordinates. 
(c) Interpreting complex charts and tables, which require detailed analysis and extraction of specific information.
}
\label{fig:high_score}
\end{figure*}

\subsubsection{How Does the Temperature in Softmax Affect Sampling?}

In our methodology, the softmax function is utilized to transform necessity scores into sampling probabilities. The temperature parameter $\tau$ in the softmax function plays a critical role in shaping the distribution of the sampled data.
Increasing the temperature parameter $\tau$ leads to a more uniform distribution, making the sampling process similar to random sampling. Consequently, the selected data may encompass a broader range of necessity scores, including both high and low extremes.
Conversely, decreasing $\tau$ focuses the sampling process towards the highest necessity scores. This results in data selection that is heavily weighted towards samples deemed most necessary by the necessity scoring mechanism.
We experimented with various $\tau$ settings to understand their effects on model performance. The results, presented in Table~\ref{tab:tau}, reveal that the choice of $\tau$ significantly influences the outcomes across different benchmarks.
From the table, it is evident that setting $\tau$ to 1 yields the best performance on four out of six benchmarks. This indicates a balanced approach where the model benefits from a mix of high-necessity data without entirely disregarding lower-scoring yet potentially valuable samples. Consequently, we adopt $\tau = 1$ as our default setting, ensuring an optimal balance between diversity and necessity in the sampled data.
However, for certain benchmarks like SQA and TQA, the highest performance is observed with $\tau = 10$. This suggests that for specific tasks, a more uniform distribution may still be beneficial, likely due to the broader coverage of data types and scenarios.

% In our method, we leverage the softmax method to convert the necessity score to sampling probability.
% The temperature $\tau$ of softmax will affect the data distribution of sampled data.
% A large $\tau$ will make the sampled data like sampled sampled data.
% A small $\tau$ will make the sampled data like sampled data with top necessity scores.
% Thus, we leverage different $\tau$ setting and display the results in Table~\ref{tab:tau}.
% We observe that, the $\tau$ is set to $1$, 4 benchmarks achieved best performance. Thus, we leverage $\tau = 1$ as default setting.

\subsubsection{How Does $k$ Affect Data Selection Performance?}

The group size $k$ is a critical hyperparameter that significantly influences the distribution of selected data. When the value of $k$ is very small, the data selection approach of \methodname closely approximates random selection. Conversely, when $k$ is exceedingly large, the selection process is primarily based on necessity scores, potentially disregarding the importance of diversity within the selected subset of data.
Our experimental evaluations indicate that the choice of $k$ has a pronounced impact on model performance. Table~\ref{tab:k} presents a detailed comparison of different $k$ values, highlighting the necessity of an optimal $k$ for achieving superior performance. Notably, the highest performance was not observed for either the smallest ($k=1\text{k}$) or the largest ($k=1000\text{k}$) values. Instead, intermediate values offered a more favorable balance, with $k=50\text{k}$ demonstrating the best performance across two different benchmarks.
Based on these findings, we adopt $k=50\text{k}$ as the default setting for \methodname. 
This value effectively balances the need for necessity and diversity, resulting in consistent performance improvements across various benchmarks.

% The group size $k$ will also affect the distribution of selected data. For example, when $k$ is very small, \methodname will be like a random selection method. when $k$ is very large, \methodname will be like only based on necessity and ignore diversity.
% Thus, a great $k$ value will make great sense. In this section, we explore the effect of different $k$ and conduct experiments by using different $k$.
% As shown in Table~\ref{tab:k}, we observe that a very large $k$ or a very small $k$ leads to Relatively poor performance. For example, The highest performance of no benchmark is achieved when $k=1\text{k}$ or $k=1000\text{k}$. Based on the experimental results, we adopt $k=50\text{k}$ as our default setting because the best performance of two benchmarks are achieved when $k=50\text{k}$.

\subsubsection{How to Allocate the Seed Data and Necessity Data?}

In our proposed \methodname, the training data is divided into two categories: seed data and necessity data. The seed data is used to train the initial seed model, which in turn generates necessity scores used to sample the necessity data.
A larger volume of seed data can potentially ensure more reliable necessity scores because it allows the model to learn better representations. However, increasing the amount of seed data necessarily decreases the amount of necessity data, due to the fixed total dataset size.
To explore the optimal balance between seed data and necessity data, we conducted experiments with different allocations. The results, displayed in Table~\ref{tab:distribution}, indicate that neither extremely large nor extremely small amounts of seed data result in optimal performance. A minimal amount of seed data leads to unreliable necessity scores, while an excessive amount of seed data leaves too little room for necessity data.
Our findings show that a configuration of 100k seed data and 565k necessity data achieves top performance in half of the benchmarks. Therefore, we adopt this allocation as our default setting. 

% In our proposed \methodname, the training data can be divided into two categories: seed data and necessity data.
% The seed data is leveraged to train the seed model.
% The necessity data is sampled based the necessity score, which is computed based on the sedd model.
% Thus, more seed data can guarantee a more reliable necessity score. However, To ensure a fixed amount of data in the training dataset, more seed data leads to less necessity data.
% To explore the balance of seed data and necessity data, we conduct experiments by setting different amounts of seed data and necessity data.
% As shown in Table~\ref{tab:distribution}, we observe that a large amount of seed data and a small amount of seed data lead to suboptimal performance. This maybe because that a small amount of seed data lead to unreliable necessity scores and a large amount of seed data lead to too small amount of necessity data.
% We observe that half benchmarks achieves best performance when setting seed data mount to 100k and necessity data to 565k, so we leverage this as default setting.

\subsubsection{What is the Impact of Different LLM Sizes on the Results?}

To evaluate the effectiveness of our proposed data selection methods across different LLM sizes, we conducted experiments using two popular configurations: Vicuna-7B and Vicuna-13B. As shown in Table~\ref{tab:size}, the performance trends are consistent regardless of the LLM size used.
Our findings indicate that the \methodname consistently outperforms both the Artificial and Random data construction methods across all benchmarks for both LLM sizes. This consistency reinforces the robustness and effectiveness of our proposed method in enhancing model performance.
Notably, while the larger Vicuna-13B model demonstrates superior performance in absolute terms, the relative improvements gained through our \methodname are similar for both Vicuna-7B and Vicuna-13B. This suggests that the benefits of our data selection approach are scalable and applicable across different model sizes.

% To explore the effectiveness of our proposed data selection methods on different LLM sizes, we conduct experiments on two popular LLM sizes, \emph{i.e.,} 7B and 13B.
% As shown Table~\ref{tab:size}, we observe that whether using Vicuna-7B or Vicuna-13B, the performance change is similar.

\subsection{Qualitative Analysis}

In \methodname, the necessity score plays a crucial role in guiding data selection. To investigate the differences in samples with varying necessity scores, we conducted an illustrative qualitative analysis.
Figure~\ref{fig:low_score} categorizes samples with low necessity scores into three main types:
\begin{itemize}
\item \textit{Easily solvable problems due to noticeable differences in options}: For instance, in the first example of Figure~\ref{fig:low_score}(a), the options are ``A. fries", ``B. cactus", ``C. onion rings", and ``D. watermelon". The vast differences between these options make it unnecessary for the model to exhibit deep understanding or reasoning about the image to select the correct answer.
\item \textit{Problems loosely related to the image, resulting in straightforward answers}: For example, in the first example of Figure~\ref{fig:low_score}(b), the question ``How many elephants are in this photo?" pertains to an image devoid of elephants. Thus, the model does not need to deeply comprehend the image to respond correctly.
\item \textit{Overly simple questions about charts}: As shown in the first example of Figure~\ref{fig:low_score}(c), the question ``Is Medium Blue the minimum? Answer yes or no." requires the model to merely identify if ``Medium Blue" represents the smallest quantity, without understanding other elements within the image.
\end{itemize}

In contrast, Figure~\ref{fig:high_score} highlights samples with high necessity scores, also categorized into three distinct types:
\begin{itemize}
\item \textit{Fine-grained text recognition problems}: For instance, in the first example of Figure~\ref{fig:high_score}(a), the question ``What is the name of the advertiser?" necessitates the model to accurately recognize all text in the image and identify the specific text corresponding to the advertiser.
\item \textit{Understanding specific regions within an image}: As depicted in the first example of Figure~\ref{fig:high_score}(b), the question ``Please provide a short description for this region: [0.74, 0.38, 1.0, 0.83]" is more challenging than a general request to describe the entire image. The model must comprehend the content of the image and accurately interpret the specified coordinate area.
\item \textit{Interpreting complex charts and tables}: An example is shown in Figure~\ref{fig:high_score}(c), where the question ``Who currently holds the positions that Matthew Lyon (dr) and Anthony New (dr) once held?" requires the model to process and understand intricate details within the table.
\end{itemize}

Our qualitative analysis reveals that samples with high necessity scores are significantly more valuable for enhancing model comprehension of complex problems. Hence, it is judicious for \methodname to prioritize data selection based on necessity scores. Nevertheless, relying exclusively on high necessity score samples can pose challenges. These samples are inherently tough, potentially complicating model learning. Moreover, this exclusive focus may overlook diversity, which is crucial for robust model performance. Therefore, incorporating a grouping strategy based on necessity scores before sampling can further optimize performance.

% 可视化不同necessity score 的样本

\section{Conclusion}\label{sec:conclusion}

This paper introduced \methodname, a novel method for automating the selection of high-quality data for Visual Instruction Tuning (VIT) by balancing necessity and diversity. Our approach leverages necessity scores and a grouped sampling strategy to ensure a comprehensive and effective dataset for training multi-modal large language models (MLLMs).
Experimental results demonstrated that \methodname significantly outperforms baseline methods like LLaVA-1.5. Remarkably, \methodname achieved superior performance with less than 1\% of the data on some benchmarks and less than 50\% on all validated benchmarks. With the same amount of data, it yielded notable improvements such as +14.54\% on DOCVQA and +25.36\% on ChartQA.

\backmatter

\bibliography{sn-bibliography}% common bib file

% Insert blank pages with a large vspace
\twocolumn[{

\clearpage
\thispagestyle{empty}
\vspace*{\fill}
\null
\vfill

\clearpage
\thispagestyle{empty}
\vspace*{\fill}
\null
\vfill

\clearpage
}]

\begin{appendices}

\section{Appendix}\label{appendix}

% An appendix contains supplementary information that is not an essential part of the text itself but which may be helpful in providing a more comprehensive understanding of the research problem or it is information that is too cumbersome to be included in the body of the paper.

\subsection{Composition of the Data Pool}\label{sec:data_pool}

We collected existing datasets and combined them into a large visual instruction tuning dataset covering a wide range of tasks including general visual question answering, captioning, OCR, document understanding, chart/figure understanding, table understanding, reasoning, logic, maths, textbook/academic questions, differences between two images, and screenshot to code, among others. The details of the data pool are presented in Table~\ref{tab:dataset_statistics_part1} and Table~\ref{tab:dataset_statistics_part2}.

For general visual question answering, we included datasets such as VQAv2 \cite{Goyal_Khot_Agrawal_Summers-Stay_Batra_Parikh_2019}, COCO-QA \cite{Ren_Kiros_Zemel_2015}, Visual7W \cite{Zhu_Groth_Bernstein_Fei-Fei_2016}, A-OKVQA \cite{Schwenk_Khandelwal_Clark_Marino_Mottaghi_2022}, TallyQA \cite{Acharya_Kafle_Kanan_2019}, OK-VQA \cite{Marino_Rastegari_Farhadi_Mottaghi_2019}, HatefulMemes \cite{Kiela_Hamed_Mohan_Goswami_Singh_Ringshia_Testuggine_2020}, and VQA-RAD \cite{Lau_Gayen_Ben_Abacha_Demner-Fushman_2018}. These datasets collectively provide tens of thousands of images, hundreds of thousands of question/answer pairs, and millions of tokens.

In the captioning task, we incorporated datasets like LNarratives \cite{pont2020connecting}, Screen2Words \cite{wang2021screen2words}, and VSR \cite{liu2023visual}, contributing a substantial number of images with corresponding textual descriptions, collectively amounting to millions of tokens.

For OCR, document understanding, and text transcription, our datasets include RenderedText \cite{RenderedText}, DocVQA \cite{mathew2021docvqa}, TextCaps \cite{Sidorov_Hu_Rohrbach_Singh_2020}, TextVQA \cite{Singh_Natarajan_Shah_Jiang_Chen_Batra_Parikh_Rohrbach_2019}, ST-VQA \cite{biten2019scene}, OCR-VQA \cite{Mishra_Shekhar_Singh_Chakraborty_2019}, VisualMRC \cite{Tanaka_Nishida_Yoshida_2022}, IAM \cite{Marti_Bunke_2002}, InfoVQA \cite{mathew2022infographicvqa}, and Diagram image-to-text \cite{DiagramImageToText}. These datasets focus on converting textual information from various document types into digital text.

For chart/figure understanding, we included datasets such as Chart2Text \cite{obeid2020chart}, DVQA \cite{kafle2018dvqa}, VisText \cite{Tang_Boggust_Satyanarayan_2023}, ChartQA \cite{masry2022chartqa}, PlotQA \cite{Methani_Ganguly_Khapra_Kumar_2019}, FigureQA \cite{Kahou_Michalski_Atkinson_Trischler_Bengio_2017}, and MapQA \cite{Chang_Palzer_Li_Fosler_Lussier_Xiao_2022}. These datasets provide a varied range of images, extensive Q/A pairs, and millions of tokens, focusing on chart and figure data.

For table understanding, we included datasets like TabMWP \cite{lu2022dynamic}, TAT-QA \cite{zhu2021tat}, HiTab \cite{Cheng_Dong_Wang_Jia_Guo_Gao_Han_Lou_Zhang_2022}, MultiHiertt \cite{zhao2022multihiertt}, FinQA \cite{Chen_Chen_Smiley_Shah_Borova_Langdon_Moussa_Beane_Huang_Routledge}, WikiSQL \cite{Zhong_Xiong_Socher_2017}, SQA \cite{Iyyer_Yih_Chang_2017}, and WTQ \cite{pasupat2015compositional}, contributing significantly in terms of tables, related Q/A pairs, and millions of tokens.

For reasoning, logic, and maths, our datasets include GeomVerse \cite{kazemi2023geomverse}, CLEVR-Math \cite{lindstrom2022clevr}, CLEVR \cite{Johnson_Hariharan_van}, IconQA \cite{lu2021iconqa}, RAVEN \cite{Zhang_Gao_Jia_Zhu_Zhu_2019}, and Inter-GPs \cite{lu2021inter}. These datasets are invaluable for tasks involving complex logical and mathematical reasoning.

For textbook and academic questions, we incorporated datasets such as AI2D \cite{kembhavi2016diagram}, TQA \cite{Kembhavi_Seo_Schwenk_Choi_Farhadi_Hajishirzi_2017}, and ScienceQA \cite{lu2022learn}. These datasets focus on educational materials and contribute a substantial number of images, Q/A pairs, and tokens pertinent to academic content.

For tasks involving differences between two images, we included datasets such as NLVR2 \cite{suhr2018corpus}, GSD \cite{li2023mimic}, and Spot the Difference \cite{jhamtani2018learning}. These datasets are designed to test the ability to identify discrepancies between pairs of images, providing thousands of image pairs and related Q/A pairs.

For the screenshot-to-code task, our dataset pool comprises WebSight \cite{laurenccon2024unlocking} and DaTikz \cite{belouadi2023automatikz}. These datasets convert screenshots into code, contributing an extensive number of tokens due to the nature of the programming languages involved.

Lastly, we included text-only data from OpenHermes-2.5 \cite{OpenHermes2.5}, which offers a sizeable collection of Q/A pairs and tokens, focusing purely on textual data without the image component.

Overall, our data pool integrates a diverse array of datasets, collectively contributing to a comprehensive and large-scale visual instruction-tuning dataset suitable for a variety of tasks. This extensive collection ensures broad coverage of different domains and challenges, thereby supporting robust model training and evaluation across multiple applications in visual and textual understanding.

\begin{table*}[]
    \caption{Composition of the Data Pool (Part 1).}
    \centering
    \setlength{\tabcolsep}{15pt}
    \begin{tabular}{l|r|r|r}
        \hline
        \hline
        \textbf{Dataset} & \textbf{\# images} & \textbf{\# Q/A pairs} & \textbf{\# tokens} \\
        \hline
        \hline
        \multicolumn{4}{l}{\textbf{General visual question answering}} \\
        VQAv2~\cite{Goyal_Khot_Agrawal_Summers-Stay_Batra_Parikh_2019} & 82,772 & 443,757 & 1,595,929 \\
        COCO-QA~\cite{Ren_Kiros_Zemel_2015} & 46,287 & 78,736 & 286,982 \\
        Visual7W~\cite{Zhu_Groth_Bernstein_Fei-Fei_2016} & 14,366 & 69,817 & 279,268 \\
        A-OKVQA~\cite{Schwenk_Khandelwal_Clark_Marino_Mottaghi_2022} & 16,539 & 17,056 & 236,492 \\
        TallyQA~\cite{Acharya_Kafle_Kanan_2019} & 98,680 & 183,986 & 738,254 \\
        OK-VQA~\cite{Marino_Rastegari_Farhadi_Mottaghi_2019} & 8,998 & 9,009 & 38,853 \\
        HatefulMemes~\cite{Kiela_Hamed_Mohan_Goswami_Singh_Ringshia_Testuggine_2020} & 8,500 & 8,500 & 25,500 \\
        VQA-RAD~\cite{Lau_Gayen_Ben_Abacha_Demner-Fushman_2018} & 313 & 1,793 & 8,418 \\
        \hline
        \multicolumn{4}{l}{\textbf{Captioning}} \\
        LNarratives~\cite{pont2020connecting} & 507,444 & 507,444 & 21,328,731 \\
        Screen2Words~\cite{wang2021screen2words} & 15,730 & 15,743 & 143,103 \\
        VSR~\cite{liu2023visual} & 2,157 & 3,354 & 10,062 \\
        \hline
        \multicolumn{4}{l}{\textbf{OCR, document understanding, text transcription}} \\
        RenderedText~\cite{RenderedText} & 999,000 & 999,000 & 27,207,774 \\
        DocVQA~\cite{mathew2021docvqa} & 10,189 & 39,463 & 337,829 \\
        TextCaps~\cite{Sidorov_Hu_Rohrbach_Singh_2020} & 21,953 & 21,953 & 389,658 \\
        TextVQA~\cite{singh2019towards} & 21,953 & 34,602 & 181,918 \\
        ST-VQA~\cite{biten2019scene} & 17,247 & 23,121 & 127,846 \\
        OCR-VQA~\cite{Mishra_Shekhar_Singh_Chakraborty_2019} & 165,746 & 801,579 & 6,073,824 \\
        VisualMRC~\cite{Tanaka_Nishida_Yoshida_2022} & 3,027 & 11,988 & 168,828 \\
        IAM~\cite{Marti_Bunke_2002} & 5,663 & 5,663 & 144,216 \\
        InfoVQA~\cite{mathew2022infographicvqa} & 2,118 & 10,074 & 61,048 \\
        Diagram image-to-text~\cite{DiagramImageToText} & 300 & 300 & 22,196 \\
        \hline
        \multicolumn{4}{l}{\textbf{Chart/figure understanding}} \\
        Chart2Text~\cite{obeid2020chart} & 26,985 & 30,242 & 2,852,827 \\
        DVQA~\cite{kafle2018dvqa} & 200,000 & 2,325,316 & 8,346,234 \\
        VisText~\cite{Tang_Boggust_Satyanarayan_2023} & 7,057 & 9,969 & 1,245,485 \\
        ChartQA~\cite{masry2022chartqa} & 18,271 & 28,299 & 185,835 \\
        PlotQA~\cite{Methani_Ganguly_Khapra_Kumar_2019} & 157,070 & 20,249,479 & 847,829,278 \\
        FigureQA~\cite{Kahou_Michalski_Atkinson_Trischler_Bengio_2017} & 100,000 & 1,327,368 & 3,982,104 \\
        MapQA~\cite{Chang_Palzer_Li_Fosler_Lussier_Xiao_2022} & 37,417 & 483,416 & 6,470,485 \\
        \hline
        \hline
    \end{tabular}
    \label{tab:dataset_statistics_part1}
\end{table*}

\begin{table*}[]
    \caption{Composition of the Data Pool (Part 2).}
    \centering
    \setlength{\tabcolsep}{15pt}
    \begin{tabular}{l|r|r|r}
        \hline
        \hline
        \textbf{Dataset} & \textbf{\# images} & \textbf{\# Q/A pairs} & \textbf{\# tokens} \\
        \hline
        \hline
        \multicolumn{4}{l}{\textbf{Table understanding}} \\
        TabMWP~\cite{lu2022dynamic} & 22,729 & 23,059 & 1,948,166 \\
        TAT-QA~\cite{zhu2021tat} & 2,199 & 13,215 & 283,776 \\
        HiTab~\cite{Cheng_Dong_Wang_Jia_Guo_Gao_Han_Lou_Zhang_2022} & 2,500 & 7,782 & 351,299 \\
        MultiHiertt~\cite{zhao2022multihiertt} & 7,619 & 7,830 & 267,615 \\
        FinQA~\cite{Chen_Chen_Smiley_Shah_Borova_Langdon_Moussa_Beane_Huang_Routledge} & 5,276 & 6,251 & 242,561 \\
        WikiSQL~\cite{Zhong_Xiong_Socher_2017} & 74,989 & 86,202 & 9,680,673 \\
        SQA~\cite{Iyyer_Yih_Chang_2017} & 8,514 & 34,141 & 1,894,824 \\
        WTQ~\cite{pasupat2015compositional} & 38,246 & 44,096 & 6,677,013 \\
        \hline
        \multicolumn{4}{l}{\textbf{Reasoning, logic, maths}} \\
        GeomVerse~\cite{kazemi2023geomverse} & 9,303 & 9,339 & 2,489,459 \\
        CLEVR-Math~\cite{lindstrom2022clevr} & 70,000 & 788,650 & 3,184,656 \\
        CLEVR~\cite{Johnson_Hariharan_van} & 70,000 & 699,989 & 2,396,781 \\
        IconQA~\cite{lu2021iconqa} & 27,315 & 29,859 & 112,969 \\
        RAVEN~\cite{Zhang_Gao_Jia_Zhu_Zhu_2019} & 42,000 & 42,000 & 105,081 \\
        Inter-GPs~\cite{lu2021inter} & 1,451 & 2,101 & 8,404 \\
        \hline
        \multicolumn{4}{l}{\textbf{Textbook/academic questions}} \\
        AI2D~\cite{kembhavi2016diagram} & 3,099 & 9,708 & 38,832 \\
        TQA~\cite{Kembhavi_Seo_Schwenk_Choi_Farhadi_Hajishirzi_2017} & 1,496 & 6,501 & 26,004 \\
        ScienceQA~\cite{lu2022learn} & 4,985 & 6,218 & 24,872 \\
        \hline
        \multicolumn{4}{l}{\textbf{Differences between 2 images}} \\
        NLVR2~\cite{suhr2018corpus} & 50,426 & 86,373 & 259,119 \\
        GSD~\cite{li2023mimic} & 70,939 & 141,869 & 4,637,229 \\
        Spot the diff~\cite{jhamtani2018learning} & 8,566 & 9,524 & 221,477 \\
        \hline
        \multicolumn{4}{l}{\textbf{Screenshot to code}} \\
        WebSight~\cite{laurenccon2024unlocking} & 500,000 & 500,000 & 276,743,299 \\
        DaTikz~\cite{belouadi2023automatikz} & 47,974 & 48,296 & 59,556,252 \\
        \hline
        \multicolumn{4}{l}{\textbf{Text-only data}} \\
        OpenHermes-2.5~\cite{OpenHermes2.5} & 0 & 1,006,223 & 248,553,747 \\
        \hline
        \hline
    \end{tabular}
    \label{tab:dataset_statistics_part2}
\end{table*}




\end{appendices}



\end{document}